\documentclass[letterpaper, 10 pt, conference]{ieeeconf}  

\IEEEoverridecommandlockouts                              

\usepackage{bm}  
\usepackage{graphics} 
\usepackage{epsfig} 
\usepackage{mathptmx} 
\usepackage{times} 
\usepackage{amsmath} 
\usepackage{amssymb}  
\usepackage{color}
\usepackage{booktabs}
\usepackage{caption}
\usepackage{subcaption}
\usepackage{float}
\usepackage[ruled,linesnumbered]{algorithm2e}
\usepackage{algorithmic}
\newcommand{\norm}[1]{\left\lVert#1\right\rVert}
\usepackage[bookmarks=false]{hyperref}
\usepackage[noadjust]{cite}

\newcommand{\RV}[1]{{\color{black}#1}}

\title{\LARGE \bf Bridging the Sim-to-Real Gap with Dynamic Compliance Tuning for Industrial Insertion
}

\author{Xiang Zhang$^{1^*}$, Masayoshi Tomizuka$^{1}$, and Hui Li$^{2}$
\thanks{$^{1}$Mechanical Systems Control Lab, UC Berkeley, USA.
        {\tt\small \{xiang\_zhang\_98, tomizuka\}@berkeley.edu}}%
\thanks{$^*$ Work done during internship at Autodesk Research}
\thanks{$^{2}$ Autodesk Research, USA
        {\tt\small hui.xylo.li@autodesk.com}}%
}

\begin{document}

\maketitle
\thispagestyle{empty}
\pagestyle{empty}

\begin{abstract}

Contact-rich manipulation tasks often exhibit a large sim-to-real gap. For instance, industrial assembly tasks frequently involve tight insertions where the clearance is less than \(0.1\) mm and can even be negative when dealing with a deformable receptacle. This narrow clearance leads to complex contact dynamics that are difficult to model accurately in simulation, making it challenging to transfer simulation-learned policies to real-world robots. In this paper, we propose a novel framework for robustly learning manipulation skills for real-world tasks using simulated data only. Our framework consists of two main components: the ``Force Planner'' and the ``Gain Tuner''. The Force Planner plans both the robot motion and  desired contact force, while the Gain Tuner dynamically adjusts the compliance control gains to track the desired contact force during task execution. The key insight is that by dynamically adjusting the robot's compliance control gains during task execution, we can modulate contact force in the new environment, thereby generating trajectories similar to those trained in simulation and narrowing the sim-to-real gap. Experimental results show that our method, trained in simulation on a generic square peg-and-hole task, can generalize to a variety of real-world insertion tasks involving narrow and negative clearances, all without requiring any fine-tuning. Videos are available at \href{https://dynamic-compliance.github.io/}{https://dynamic-compliance.github.io}

\end{abstract}


\section{Introduction}
Industrial robots are increasingly deployed to address complex manipulation tasks that are contact-rich, such as parts assembly \cite{inoue2017deep,zhang2022learning,vuong2021learning, luo2019reinforcement}, and object pushing \cite{zhou2022learning,zhang2023learning,stuber2020let,hausman2018learning}. These tasks require the robot to interact with its environment by applying the appropriate contact force. However, the inherently high-dimensional and nonlinear nature of contact dynamics poses a significant challenge for automation. Traditional approaches often require human expertise for the manual design of robot skills, which is not only time-consuming but also poorly scalable across varying task settings.

Recent advancements in learning-based approaches offer a promising approach for automating the acquisition of complex robot contact-rich manipulation skills. While earlier studies concentrated on learning these skills directly on physical robot systems, the focus has recently shifted to conducting most of the learning in simulations to address the training safety and efficiency. However, accurate simulation of complex contact dynamics is still a considerable challenge since simulation is sensitive to parameters like surface friction and material stiffness \cite{parmar2021fundamental}. Moreover, task-specific geometries, such as irregular connectors, and tight peg-hole clearances, exacerbate the challenges. Consequently, the sim-to-real transfer of learned skills remains a hurdle. 


\begin{figure}
    \centering
    \includegraphics[width=240pt]{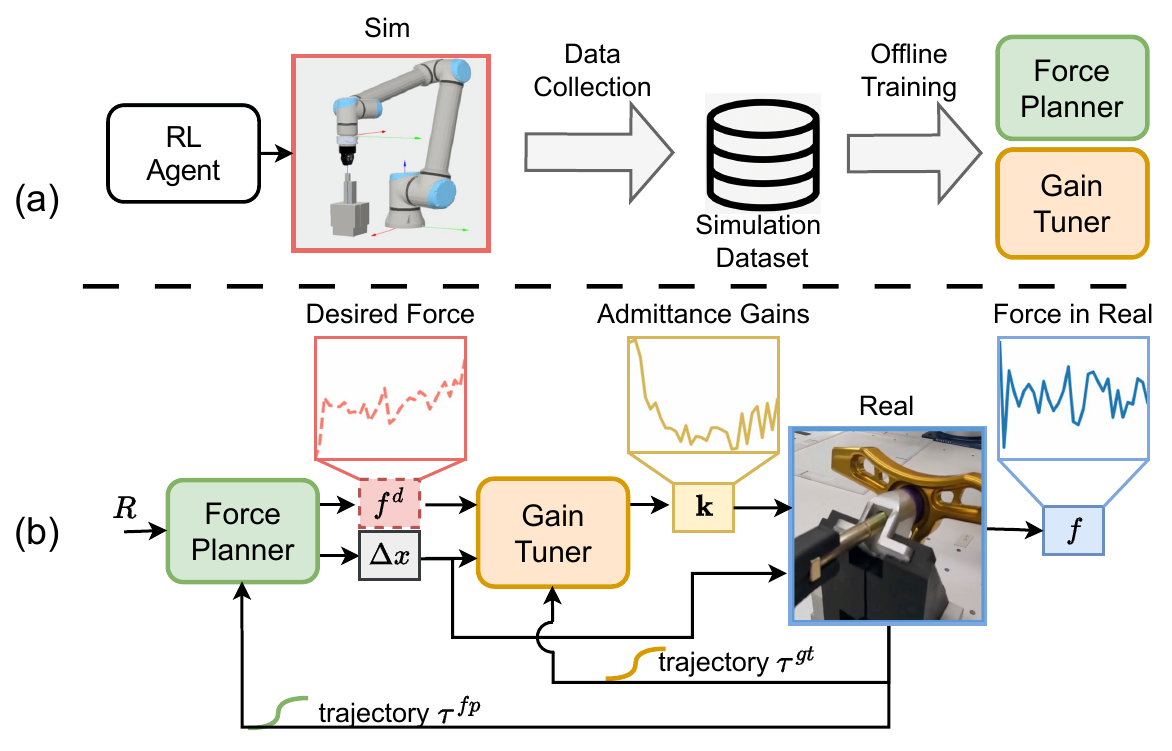}
    \caption{Overview of the proposed method. a) In simulation, we use an RL agent to generate data to offline train both the Force Planner and the Gain Tuner. b) In real-world deployment, the Force Planner plans the desired force $f^d$ and robot motion $\Delta x$ to achieve the target return $R$. The Gain Tuner then dynamically adjusts the admittance gains $\mathbf{k}$ to track the desired force.}
    \label{fig:over_view_all}
\end{figure}

Another challenge in the sim-to-real transfer of learned robot skills is compliance control. For contact-rich manipulation tasks, compliance control is often deployed to achieve safe contact during manipulation. However, the compliance control gains for real robots are difficult to tune to maintain stability, and often vary from task to task. Moreover, different phases within a single task might require variable gains for improved performance. For example, a polishing robot should be stiff in free space but compliant when making contact with a surface to avoid damage. Previous studies have proposed learning approaches to automatically obtain the compliance gains \cite{beltran2020variable,seo2023robot, buchli2011learning,wang2022safe}. Nevertheless, these learned control gains can suffer from the sim-to-real gap and require fine-tuning with real-world data. 

This paper aims to robustly train robots for real-world, contact-rich tasks using only simulated data. Our core hypothesis is that by adjusting the robot's compliance control gains during task execution, we can adaptively modulate contact force resulted in the new environment to match the desired contact force and thus generate trajectories akin to those trained in simulation. Our novel framework comprises two main components: the ``Force Planner" and the ``Gain Tuner", as illustrated in Fig.~\ref{fig:over_view_all}. The Force Planner plans the robot motion and desired contact force given the desired return and previous robot trajectory. The Gain Tuner serves as a gain scheduler, determining the most appropriate admittance control gains to track the desired contact force based on historical data. Both modules are trained exclusively in simulation in a supervised learning fashion and are directly transferable to real-world scenarios without fine-tuning. 

We validate our approach on a diverse set of challenging insertion tasks: 1) 3D-printed peg-and-holes with clearance of 0.02 mm and 0.05 mm; 2) electronic connectors, including the Ethernet and waterproof types; 3) real-world skateboard truck assembly with deformable receptacles and negative clearance. Our method demonstrates excellent zero-shot transferability, outperforming baseline approaches, and affirming its effectiveness and robustness.
\section{Related Work}
\subsection{Learning-Based Contact-Rich Manipulation}
Recent research focuses on leveraging learning-based approaches to automatically acquire contact-rich manipulation skills. These methods include imitation learning from expert demonstrations \cite{peternel2015human,abu2018force,zhang2021learning,rey2018learning,seo2023robot, wu2022prim}, as well as reinforcement learning (RL) through trial and error \cite{buchli2011learning,inoue2017deep,zhang2022learning, vuong2021learning, luo2019reinforcement, zhou2022learning,zhang2023learning,beltran2020variable,martin2019variable,chitnis2020efficient,tang2023industreal,handa2023dextreme,andrychowicz2020learning,schoettler2020meta,luo2020dynamic,spector2020deep,luo2021robotless,wu2023zero}. One line of work in this area focuses on directly learning manipulation policies on real robots. In these cases, training efficiency and safety are major concerns and have been explored in various studies \cite{inoue2017deep,luo2019reinforcement,buchli2011learning,rey2018learning,chitnis2020efficient,insertionnet,insertionnet2.0,fu2023safe}. Another line of research aims to acquire manipulation skills in simulation and subsequently transfer them to the real world \cite{zhang2022learning,vuong2021learning,zhou2022learning,zhang2023learning,beltran2020variable,tang2023industreal,martin2019variable}. However, discrepancies between simulated and real-world contact dynamics create significant obstacles for transferring learned skills. To address this issue, several approaches have been proposed. Domain randomization methods have been utilized, as discussed in \cite{zhang2023learning,zhou2022learning,beltran2020variable,chebotar2019closing,handa2023dextreme,andrychowicz2020learning}, to broaden the simulation environments in order to better capture real-world scenarios. Additionally, system identification techniques \cite{ljung1998system,Lim2022Real} actively calibrate simulation parameters to match real-world trajectories. Authors in \cite{tang2023industreal} improve simulation accuracy during policy learning by assigning low weights to simulation trajectories with large interpenetration. Apart from purely simulation-based methods, fine-tuning the learned skill with a small amount of real-world data is also a common practice in manipulation tasks \cite{schoettler2020meta,beltran2020variable}. In this work, we only use simulation data and bridge the sim-to-real gap by dynamic adjustment of compliance control gains. 

\subsection{Robot Force Control for Manipulation}
A suitable force controller is needed to regulate the contact force during manipulation.
Pure force control methods are not suitable for manipulation tasks as they require a specific desired force input and are ineffective in free space \cite{song2017impedance}. Position-force control methods are more common and fall into two categories: hybrid position-force control and impedance/admittance control. Hybrid position-force control employs two separate feedback laws—one for position and the other for force—combined using a selection matrix. This controller has been used for tasks like robotic insertion \cite{beltran2020variable,lin2016robot}, but the switching between position and force tracking can result in unstable responses. In contrast, impedance/admittance control generates force based on the deviation from the desired trajectory and does not require switching, making it a popular choice in recent research \cite{luo2019reinforcement,peternel2015human,abu2018force,zhang2021learning,buchli2011learning,rey2018learning,martin2019variable,kozlovsky2022reinforcement,seo2023robot}.

Tuning the gains of a force controller presents another challenge in contact-rich manipulation tasks. 
Some previous works have relied on pre-tuned force controllers and focused solely on learning robot motion \cite{zhang2022learning,vuong2021learning,zhou2022learning,zhang2023learning,narang2022factory,tang2023industreal}. However, force control gains for real robots
are difficult to tune to maintain stability and compliance, and often vary from task to task. \cite{martin2019variable} suggests jointly learning robot motions and variable gains, demonstrating advantages over a fixed-gain baseline. This approach has been employed in various studies \cite{peternel2015human,abu2018force,zhang2021learning,buchli2011learning,rey2018learning,martin2019variable,kozlovsky2022reinforcement,seo2023robot}, but most still require a real-world training or fine-tuning phase to obtain feasible gains for real-world applications. Moreover, direct transfer of these skills is typically limited to simple tasks like waypoint tracking and surface wiping, as indicated by \cite{martin2019variable}. The authors in \cite{zhang2023efficient} uses an online optimization method to adjust the admittance gains online to minimize a pre-defined cost. However, solving the nonconvex optimization is time consuming and their method results in a low-frequency gain adaption. In this work, we directly transfer the simulation-learned skills to real-world deployment on tight insertion tasks without fine-tuning, by predicting the desired contact force and tracking it with dynamic gain tuning. 

\section{Problem Description}

In this paper, we solve the problem of direct sim-to-real transfer of tight insertion skills. We first introduce the admittance controller for the robotic insertion tasks and then describe the problem setup.
\subsection{Admittance Control}
\begin{figure}
    \centering
    \includegraphics[width=200 pt]{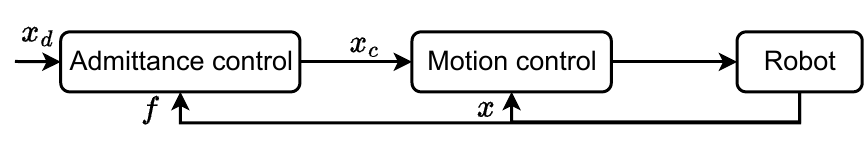}
    \caption{Admittance control loop}
    \label{fig:admittance}
\end{figure}
We utilize an admittance controller \cite{ott2010unified} to manage the robot's actions. The control framework is depicted in Fig.~\ref{fig:admittance}. The admittance controller takes in the desired pose and velocity $x_d, \dot{x}_d$ and generates compliant robot motion $x_c, \dot{x}_c$ according to a mass-spring-damping dynamics model driven by the external force $ f $:
\begin{equation}
\label{eqn_admittance_law}
M (\ddot{x}_c - \ddot{x}_d) + D (\dot{x}_c - \dot{x}_d) + K (x_c - x_d) = f
\end{equation} 
Here, $ M, K, $ and $ D $ are the inertia, stiffness, and damping matrices, respectively. We assume $ M, K, D $ are positive definite diagonal matrices with diagonal entries $ m_i, k_i, d_i > 0 $ for $ i = 1, \ldots, 6 $ to ensure stability while reducing dimensionality. The compliant robot motion $ x_c, \dot{x}_c $ is then tracked by a low-level motion controller.

\subsection{Problem Setup} \label{sec:setup}
\begin{figure}
    \centering
    \includegraphics[width=240pt]{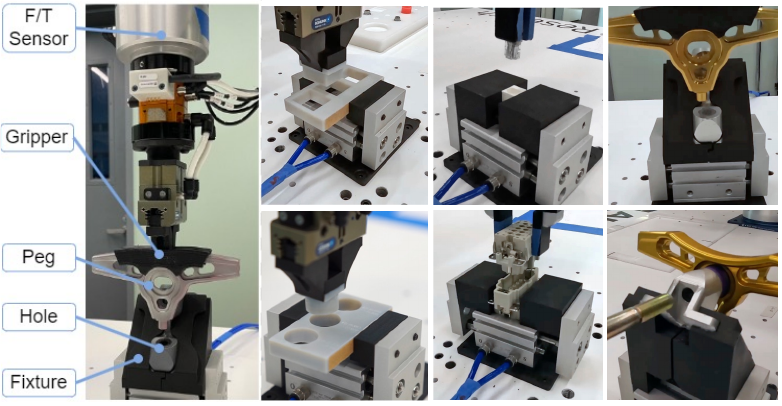}
    \caption{Real robot setup and insertion tasks}
    \label{fig:task_setup}
\end{figure}

Our real robot setup, as shown in Fig.~\ref{fig:task_setup}, consists of a UR10e robot mounted on a work table, an embedded force/torque (F/T) sensor located at the robot's wrist, a Schunk or Robotiq gripper grasping a peg and inserting it into a fixed hole. There is no camera available. During insertion, we assume that information about the nominal hole location is known but there is uncertainty regarding the exact hole pose. We rely on the F/T sensor and robot proprioception for decision-making, rather than using vision. This approach is beneficial for tight assembly tasks, where large occlusion occurs, and contact force can help align the peg and hole with high precision. 
Specifically, we solve three sets of challenging insertion tasks:
\begin{itemize}
   \item Tight 3D-printed rectangular and round peg-and-holes, each with two levels of clearance: $0.05~mm$ and $0.02~mm$, which are an order of magnitude smaller than in previous work.
   \item Complex-shaped electronic connectors, including standard Ethernet and waterproof connectors.
   \item Real-world skateboard truck assembly. The first task is inserting the hanger into a deformable receptacle, and the second task is inserting a kingpin with threads on the tip into the truck base while passing through a tight rubber bushing. Both tasks have negative clearance.
\end{itemize}

We model the robot insertion task as a Markov Decision Process (MDP), denoted as $ \{S, A, R, P, \gamma \} $, where $ S \in \mathbb{R}^{18} $ contains the peg pose $ x \in \mathbb{R}^6 $, peg velocity $ \dot{x} \in \mathbb{R}^6 $, and contact force $ f \in \mathbb{R}^6 $. The action space $ A \in \mathbb{R}^{12} $ includes the incremental robot Cartesian motion $ \Delta x \in \mathbb{R}^6 $ and the diagonal entries of the stiffness matrix $ \mathbf{k} = \{ k_1, \dots, k_6 \} $. The reward function $ R $ is defined as $ r = -||x_{\text{pos}} - x_d||_2 $, which penalizes the Euclidean distance between the robot's current position and a fixed target point inside the hole. $ P $ represents the state-transition probability, and $ \gamma $ is the discount factor. It is important to note that we simplify the task by planning only for the robot stiffness $ K $, while keeping the inertia matrix $ M $ fixed. We compute the damping matrix $ D $ as $ D = 4\sqrt{MK} $ to ensure an overdamped system. This setup is commonly employed, as referenced in relevant works \cite{abu2018force,seo2023robot}. We use the admittance gain to refer to the diagonal entries $\mathbf{k}$ of the stiffness matrix in the rest of the paper for simplicity.

\section{Proposed Approach}

The transfer from simulation to the real world is particularly challenging for contact-rich tasks, as the same robot motion and force control gains that are effective in simulation can lead to different contact force in the real world. To tackle this issue, we introduce a learning framework comprising two main modules: 1) the \emph{Gain Tuner} that generates appropriate force control gains based on planned robot motion, desired contact force, and historical trajectory data; and 2) the \emph{Force Planner} that plans the robot motion and desired contact force required to complete the task. 

The Gain Tuner serves as a novel component designed to adjust force control gains during execution to meet the desired contact force, thus bridging the sim-to-real gap. This is an improvement over baseline methods, which commonly generate both robot motion and control gains simultaneously. Both the Gain Tuner and the Force Planner are trained using simulation data only, enhanced with domain randomization and data augmentation techniques to boost robustness. Details on the model architectures and data collection processes are provided in the subsequent sections.

\subsection{Gain Tuner} \label{sec:GT}

During manipulation, the next contact force $ f_{t+1} $ depends on the current state $ s_t $, the robot motion $ \Delta x_t $, admittance gain $ \mathbf{k}_t $, and environmental properties such as part geometry, friction, and surface stiffness, denoted by $ E $. We model this relationship as a probability distribution $ P(f_{t+1} | \Delta x_t, \mathbf{k}_t, s_t, E) $. Previous approaches directly output $ \Delta x_t $ and $ \mathbf{k}_t $ from the policy; however, the contact force $ f_{t+1} $ varies during sim-to-real transfer because $ E $ changes. To address this issue, we decompose the probability as follows:
\begin{equation}
    P(f_{t+1}|\Delta x_t, \mathbf{k}_t, s_t, E) = \frac{P(\mathbf{k}_t|f_{t+1},\Delta x_t,s_t,E)P(\Delta x_t, f_{t+1}|s_t,E)}{P(\Delta x_t, \mathbf{k}_t|s_t, E)}
\end{equation}
Given the distribution of the robot motion and next contact force $ P(\Delta x_t, f_{t+1}|s_t,E) $, we can adjust the admittance gain $ \mathbf{k}_t $ relative to $ \Delta x_t, s_t $, and $ E $ to align the contact force achieved by the robot $ P(f_{t+1}|\Delta x_t, \mathbf{k}_t, s_t, E) $ with the target distribution.

In practice, the distribution $ P(\Delta x_t, f_{t+1}|s_t,E) $ is often unknown. We thus directly learn a Force Planner model as described in Sec.~\ref{sec:FP}. We propose a Gain Tuner model, $ GT(\mathbf{k}_t|f^d_{t+1},\Delta x_t,s_t,E) $, that tunes the admittance gain automatically to match the actual force with the desired force $f^d$. Modeling environmental properties $ E $ is challenging, but inspired by recent works \cite{chen2021decision,janner2021offline} that treat RL as a sequence modeling problem, we use previous trajectory data $ \tau^{gt} $ to approximate $ E $:
\begin{align}
\begin{split}
    \tau^{gt}_t = (x_{t-H}, \dot{x}_{t-H}, \Delta x_{t-H}, \mathbf{k}_{t-H}, f^d_{t-H+1}, \\
    \dots, x_{t-1}, \dot{x}_{t-1}, \Delta x_{t-1}, \mathbf{k}_{t-1}, f^d_{t})
\end{split}
\label{eqn:tau_gt}
\end{align}
where $ H $ is the window size. Our intuition is that the environment $ E $ is encoded in the previous trajectory $ \tau^{gt} $, and therefore we can infer $ E $ from $ \tau^{gt} $. The overall Gain Tuner model is  $GT(\mathbf{k}_t|f^d_{t+1},\Delta x_t,x_t,\dot{x}_t,\tau^{gt}_t)$. At each time step $ t $, the inputs for the Gain Tuner are $ \tau^{gt}_t, x_t, \dot{x}_t, \Delta x_t $, and the planned next force $ f^d_{t+1} $. The output is the predicted admittance gains $ \mathbf{k}_t $. During training, we replace the planned next force $ f^d_{t+1} $ with the ground truth $ f_{t+1} $ available from the dataset, and the training loss is the mean-squared error (MSE) between the predicted and actual admittance gains.

\subsection{Force Planner}\label{sec:FP}

In our framework, the Force Planner is designed to generate both the desired robot motion and the corresponding contact force. We employ the Decision Transformer (DT)~\cite{chen2021decision} to train the Force Planner using an offline dataset, collected according to Sec.~\ref{sec:data}. \RV{DT uses the GPT architecture to autoregressively model trajectories and predicts the next action based on historical data.} To facilitate training, we introduce an extended action space $ A' = [\Delta x, f^d] $, which includes robot motion $ \Delta x $ and desired next contact force $ f^d $.

For each training iteration, we first sample a historical trajectory of a window size $ H $ from the dataset. This trajectory is denoted as:
\begin{align}
    \begin{split}
        \tau^{fp}_t = &(s_{t-H}, a'_{t-H}, R_{t-H}, \ldots, s_{t-1}, a'_{t-1}, R_{t-1})\\
        = &(x_{t-H},\dot{x}_{t-H},f_{t-H}, \Delta x_{t-H}, f^d_{t-H+1},R_{t-H}, \dots,\\
        &x_{t-1},\dot{x}_{t-1},f_{t-1}, \Delta x_{t-1}, f^d_{t},R_{t-1})
    \end{split}
    \label{eqn:tau_fp}
\end{align}
Where $ R_t = \sum_{t'=t}^{T} r_{t'} $ is the desired future return until the last timestep of the trajectory $ T $.
Then, $ \tau^{fp}_t $ is combined with the current robot state $ s_t = [x_t,\dot{x}_t,f_t] $ and the desired return $ R_t $ to serve as the input for the Force Planner $ FP(\Delta x_t, f^d_{t+1}|x_t,\dot{x}_t,f_t,R_t,\tau^{fp}_t) $. This model predicts the subsequent robot motion $ \Delta x_t $ and the next contact force $ f^d_{t+1} $.

Similar to the Gain Tuner, an MSE loss for the robot motion and next contact force is enforced to train the Force Planner. Detailed training steps can be found in Algorithm~\ref{alg_GTFP}.

\begin{algorithm}
\caption{Learning GT and FP}
\begin{algorithmic}[1] 
\label{alg_GTFP}
\REQUIRE A set of simulated trajectories $\{\tau_i| i=1,\dots,N\}$
\STATE Initialize the dataloader D = []
\FOR{i in 1,\dots,N}
    \FOR{t in 0,\dots, T-1}
        \STATE Add $[x_t,\dot{x}_t, \Delta x_t, k_t, f_{t+1}]$ to D
        \STATE Obtain $[\tau^{gt}_t,\tau^{fp}_t]$ using (\ref{eqn:tau_gt}) (\ref{eqn:tau_fp}) and add to the D
        \STATE Add $ R_t = \sum_{t'=t}^{T} r_{t'}$ to the D
    \ENDFOR
\ENDFOR
\FOR{batched $[x,\dot{x}, \Delta x, k, f, f_{next}, R,\tau^{gt},\tau^{fp}]$ in D}
\STATE $\hat{\mathbf{k}} = GT(\mathbf{k}|f_{next},\Delta x,x,\dot{x},\tau^{gt})$
\STATE $Loss_{gt} = \norm{\mathbf{k}-\hat{\mathbf{k}}}_2^2$
\STATE $\Delta\hat{x}, \hat{f}_{next} = FP(\Delta{x}, {f}_{next}|x,\dot{x}, f, R,\tau^{fp})$
\STATE $Loss_{fp} = \norm{\Delta x-\Delta\hat{x}}_2^2 + \norm{{f}_{next}-\hat{f}_{next}}_2^2$
\STATE Update GT and FP to minimize $Loss_{gt}, Loss_{fp}$
\ENDFOR

\end{algorithmic}
\end{algorithm}

\subsection{Data Collection and Augmentation} \label{sec:data}

Both the Gain Tuner and the Force Planner in our framework are trained through supervised learning using simulation data. Figure~\ref{fig:over_view_all}(a) illustrates our simulation environment, which models a square peg-and-hole with edges of $40~\text{mm}$ in the PhysX~\cite{PhysX} physics engine. 
To generate the dataset, we employ an RL agent and use a curriculum-based approach consisting of two phases. Initially, we set a larger clearance of $0.5~\text{mm}$ and utilize the Soft-Actor-Critic algorithm \cite{haarnoja2018soft} to learn an insertion policy. This policy serves as the starting point for training on a more challenging task with a $0.3~\text{mm}$ clearance. The replay buffer collected during the RL training of the second phase is used as the dataset for training both the Gain Tuner and the Force Planner networks.

To ensure the robustness of these networks, we apply domain randomization and data augmentation techniques. We apply domain randomization in the peg's initial pose and the hole location. We also augment the contact force data with scaling and noise injection. The intuition behind this augmentation comes from real-world scenarios. Since the admittance command is controlled by the low-level robot motion controller, which often exhibits different characteristics in simulation and reality, the same admittance gains can produce varying contact force. To capture this discrepancy, we introduce a scaling factor ranging from $0.4$ to $1.4$ to the contact force data. Additionally, we inject zero-mean Gaussian noise with a standard deviation of $1~N$ to model the noisy F/T sensor readings in real-world scenarios. 
\RV{We have found that randomizing dynamics properties such as mass and friction can cause unstable behaviors in simulation and therefore, only augmented the force data.}

\subsection{Training Details }

We adopt a model architecture similar to that of DT for both the Gain Tuner and the Force Planner. For the Gain Tuner, the stack of robot states and motions $(x_{t-H}, \dot{x}_{t-H}, \Delta x_{t-H}, \dots, x_t, \dot{x}_t, \Delta x_t)$, previous admittance gains $(\mathbf{k}_{t-H}, \dots, \mathbf{k}_{t-1})$, and desired force $(f^d_{t-H+1}, \dots, f^d_{t+1})$ are mapped separately to three 128-dimensional embedding spaces using linear layers. These embeddings are then processed by the GPT-2 model~\cite{gpt-2} to predict the admittance gain $\mathbf{k}_t$. The Force Planner employs the same architecture but utilizes different modalities, including the state $s$, extended action $a'$, and return $R$, to obtain the embeddings.

We use the dataset described in Sec.~\ref{sec:data} to train both networks. \RV{Window size $H$ is 20 for both models.} During each training step, we sample trajectories from the dataset using a batch size of 64 and update both networks with a learning rate of $5 \times 10^{-4}$. The training process spans a total of $200,000$ steps. \RV{The other training hyperparameters remain the same as the original DT implementation~\cite{chen2021decision}.}

\section{Experiments}

In this section, we introduce experiments that evaluate the sim-to-real robustness and generalizability of our method. We also provide comparisons with two baselines, both of which output robot motion $\Delta x$ and admittance gains $\mathbf{k}$ simultaneously: 1) a DT that is trained on the same dataset, and 2) an RL policy that is obtained during the data collection process with $0.3~mm$ clearance as described in \ref{sec:data}. We include the two baselines to show that the decoupling of admittance gains and robot motion helps to bridge the sim-to-real gap. In addition, we include the ablation study to answer two questions: 1) how does the Gain Tuner respond to different desired force? and 2) can fine-tuning the Force Planner with a small amount of real data achieve better performance for difficult tasks?
\subsection{Experimental Setup}
We use the UR10e robot with a built-in F/T sensor on the wrist for real-world experiments. Before each trial, the robot grasps the peg and the receptacle is fixed on the table. The initial position of the peg is randomly sampled within the range of half the peg size in $X,Y,Z$ axes. We evaluate our approach on a diverse set of insertion tasks including low and negative clearances, shown in Fig.~\ref{fig:task_setup}. Their respective clearance sizes are listed in Table.~\ref{table:tasks}.

\begin{table}[h!]
  \centering
  
  \begin{tabular}{cccc}
    \toprule
             & Peg Size (mm) & Hole Size  (mm)& Clearance (mm)\\
    \hline
    Rectangular & 30.00 & 30.05 & 0.05 \\
    Rect. (tight) & 30.00  & 30.02 & 0.02 \\
    Round & 24.98 & 25.03 & 0.05 \\
    Round (tight) & 24.98 & 25.00 & 0.02 \\
    Connectors & Standard  & -- & --  \\
    Hanger& 9.56 & 9.35 & -0.21 \\
    Kingpin& 9.41 & 9.20 & -0.21 \\
    \bottomrule
  \end{tabular}
  \caption{Clearance sizes of the insertion tasks, only listing the long edge of the rectangular peg-and-hole.}
  \label{table:tasks}
\end{table}
\begin{figure*}[!h]
    \centering
    \includegraphics[width=450pt]{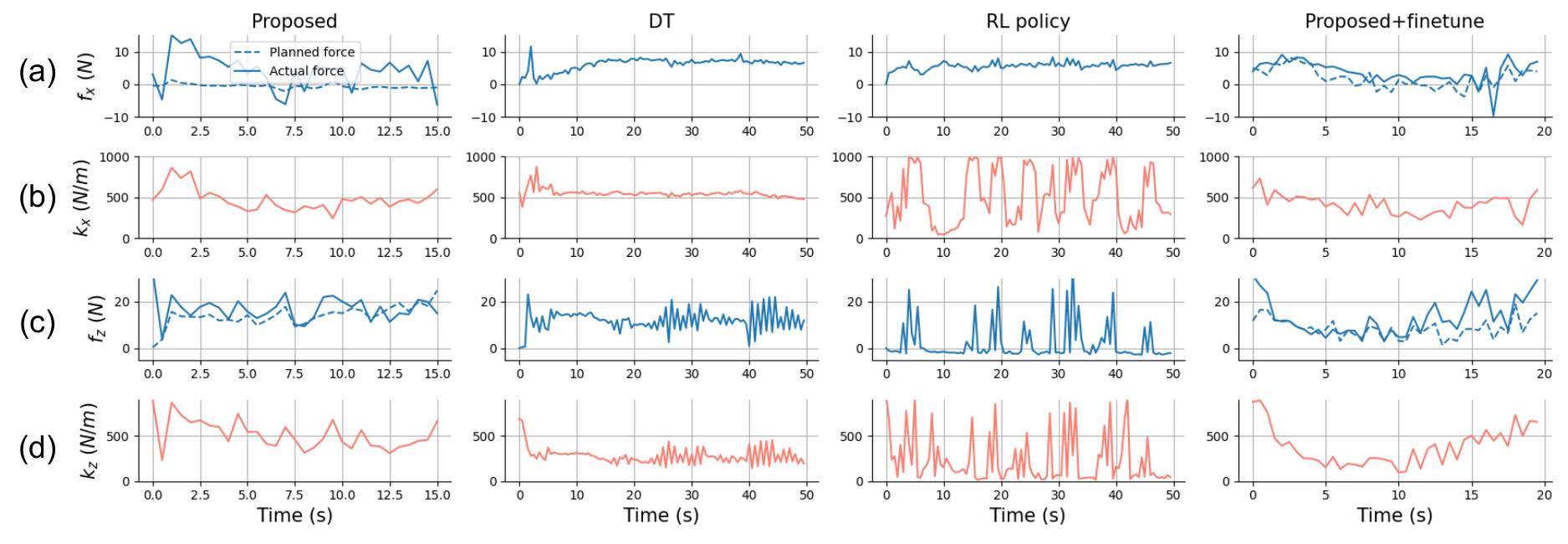}
    \caption{a) and c) show the contact force during the rectangular insertion along the $X,Z$ axes, respectively. b) and d) depict the corresponding admittance gains. Our method adaptively adjusts the gains to track the planned contact force generated by the Force Planner. Two baseline methods suffer from the sim-to-real gap and generate noisy gains and contact force.} 
    \label{fig:comparison}
\end{figure*}

\subsection{Tight 3D-Printed Peg-and-Holes}
\begin{table*}
  \centering
  \begin{tabular}{ccccccccc}
    \toprule
             & Rectangular & Rect. (tight)& Round & Round(tight)& Ethernet & Waterproof & Hanger & Kingpin\\
    \hline
    Proposed & $\bm{9/10}$ & $\bm{7/10}$ & $\bm{10/10}$ & $\bm{9/10}$& $\bm{10/10}$ & $\bm{9/10}$ & $\bm{9/10}$ & $\bm{7/10}$\\
    Decision Transformer & 1/10 & 0/10 & 4/10 & 2/10& 8/10 & 5/10 & 7/10 & 3/10\\
    RL Policy & 1/10 & 3/10 & 5/10 & 5/10& 6/10 & 6/10 & 3/10 & 5/10\\
    \bottomrule
  \end{tabular}
  \caption{Success rate comparison with baseline methods on insertion tasks. }
  \label{table:success_rates}
\end{table*}

We first conduct experiments on tight 3D-printed peg-and-holes to evaluate the sim-to-real transfer performance. The main challenge in these tasks comes from the tight clearances of $0.05~mm$ and $0.02~mm$. To complete the insertion task, the robot must align the peg and the hole with high precision, necessitating a robust search strategy and suitable force control gains. The success rates are reported in Table~\ref{table:success_rates}. Our method achieves $90\%$ and $100\%$ success rate for the rectangle and round peg-and-holes with $0.05mm$ clearance, and $70\%$ and $90\%$ success rate for those with $0.02mm$ clearance, all through zero-shot sim-to-real transfer. In contrast, both the DT and RL models struggle with sim-to-real transfer, failing most trials.

As seen in Table~\ref{table:success_rates}, the rectangular peg-in-hole tasks are particularly difficult for baseline methods. We investigate the cause by examining the contact force profile and the admittance gains across all methods in Fig.~\ref{fig:comparison}. We omit the $Y$ axis because $X$ and $Y$ are similar for the insertion tasks. In simulation, the robot does not make contact with the hole during the first step in this example. Thus, the initial planned contact force in the $Z$ axis is $0~N$ and the Gain Tuner predicts a stiffness above $800~N/m$. However, the actual initial contact force in the $Z$ axis is $33~N$, much higher than planned. The Gain Tuner then significantly reduces the stiffness to match the second planned force of $4~N$. For subsequent steps, the Force Planner continually updates the planned force while the Gain Tuner adjusts the robot's stiffness accordingly, resulting in a force tracking error of around $5~N$ in the $Z$ axis. This demonstrates the effectiveness of our Gain Tuner. 

In comparison, because the DT and RL baselines directly output both robot motion and admittance gains, they lack the capability to modulate the contact force. The DT model maintains robot stiffness below $400~N/m$ and generates noisy contact force. Meanwhile, the RL baseline alternates rapidly between high and extremely low stiffness, inducing a pulsatile pattern in the contact force. Neither baseline manages to uphold appropriate contact, leading to diminished success rates in challenging insertion tasks.

It is worth noting that our method does not align well the planned and the actual force in the $X$ axis. This discrepancy is primarily because the simulated contact force in the $X, Y$ axes are close to $0~N$, attributable to an inaccurate contact model in simulation. Additionally, our real-world tasks have smaller clearances, leading to higher sticking force between the peg and the hole surface.

\subsection{Electronic Connectors}
We then examine how our method generalizes to different geometries from in simulation by selecting the Ethernet and waterproof connectors. These were chosen due to their ubiquity in insertion tasks and their complex, non-convex geometries. Given that the clearances are larger compared to other tasks, both DT and RL achieve success rates exceeding $50\%$. However, our proposed method continues to outperform the baselines, achieving a $100\%$ success rate for the Ethernet connector and $90\%$ for the waterproof connector.

\subsection{Skateboard Truck Assembly}

We next aim to address a real-world assembly task - skateboard truck assembly. We focus on two subtasks: hanger and kingpin assembly. These tasks are challenging due to deformable parts within the holes, which create negative clearances and consequently increase the contact force. Our method achieves $90\%$ success rate in the hanger task, thereby demonstrating its generalizability. The DT and RL baselines occasionally apply excessive force, leading to deformations that cause the peg and hole to stick together and reducing their success rates. 

The kingpin insertion task involves metal-to-metal contact. Because both objects are stiff, such contact often generates a large amount of force, triggering protective stops. Our method is not immune to this issue and has a success rate of $70\%$, primarily because the Gain Tuner initially predicts high stiffness based on simulation data. However, it still outperforms both baselines.

\subsection{Ablation Studies}

\begin{figure}
    \centering
    \includegraphics[width=230pt]{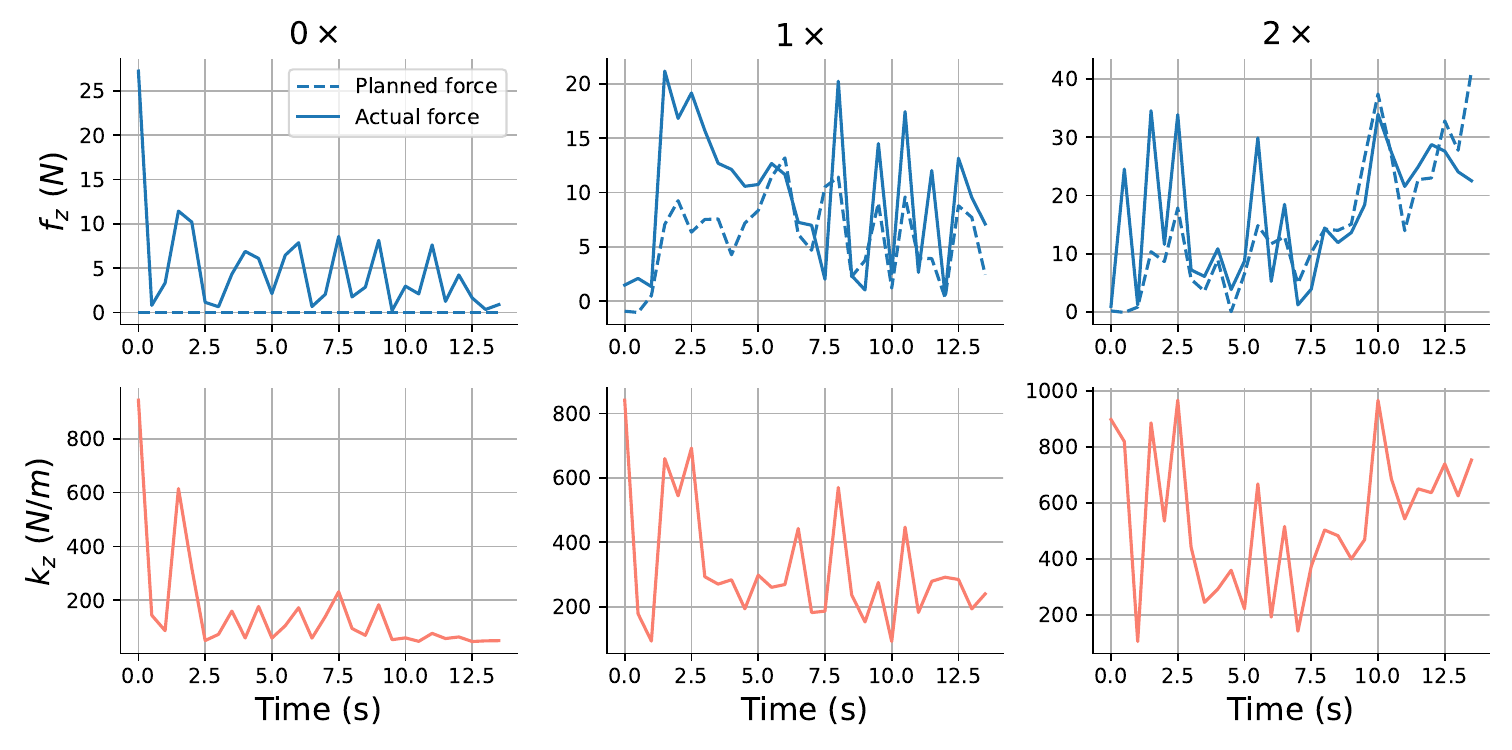}
    \caption{The contact force and admittance gains in the $Z$ axis generated by the Gain Tuner while scaling the desired force.}
    \label{fig:ablation}
\end{figure}

In this subsection, we assess the capability of the Gain Tuner to generate varying gains for different desired force. We multiply the desired force output from the Force Planner by a set of factors ranging from \(0\) to \(2\), while keeping the robot motion intact. The resulting contact force and predicted gains are depicted in Fig.~\ref{fig:ablation}. In the case where the factor is \(0\), the planned robot motion still pushes downward, resulting in a non-zero contact force in the \(Z\) axis. Nevertheless, the Gain Tuner rapidly adjusts the stiffness to below \(200 \, \text{N/m}\), and the subsequent contact forces are mostly below \(10 \, \text{N}\). These results underscore the effectiveness of our proposed method in dynamically adjusting admittance gains during deployment. As we increase the multiplication factor, the Gain Tuner correspondingly elevates the robot's admittance gain to track the planned force, as illustrated in Fig.~\ref{fig:ablation}.

We also answer the question of whether fine-tuning with a small amount of real data can achieve better performance for the difficult tasks. \RV{We hypothesize that the main simulation inaccuracy is in contact dynamics modeling and the main sim-to-real gap is in contact force prediction. Therefore, we only fine-tune the Force Planner.} We collect 10 real-world trajectories for the $0.05mm$ rectangular peg-in-hole task for fine-tuning. We evaluate the fine-tuned Force Planner with the original Gain Tuner on the difficult tasks: $0.02mm$ rectangular peg-in-hole and kingpin insertion. This ablation achieves improved performance for both, $90\%$ success rate for tight rectangular and $100\%$ for kingpin. For the force comparison, Fig.~\ref{fig:comparison} shows that this ablation yields more realistic force plans in the $X$ direction and outperforms our initial method in the rectangular task. 
\RV{This shows that fine-tuning the Force Planner can improve both the planned contact force and overall performance, while the original Gain Tuner still works well for modulating the contact force.}

\section{Conclusion and Discussions}

In this paper, we propose a novel method for bridging the sim-to-real gap in robot manipulation tasks that involve tight tolerance and complex contact dynamics. Our method comprises two main components: Gain Tuner, which adjusts the robot's admittance gains to track the desired contact force, and Force Planner, which generates the robot's desired motion and desired contact force. We train both components solely in simulation and achieve excellent zero-shot sim-to-real transfer and generalization to a diverse set of real-world tight insertion tasks. Experimental results in comparison with baseline methods demonstrate the advantage of explicit modulation of the contact force by dynamically adjusting the compliance control gains during execution.

Despite the success, our method has limitations: 1) although the desired contact force provides a good guidance to a successful insertion trajectory, due to variations in task settings, the Force Planner trained purely in simulation may not output realistic desired contact force for some real-world deployment; 2) we simplify the insertion task by assuming that the peg is pre-grasped by the robot. For future work, we aim to explore methods for efficiently adapting the Force Planner online to generate better force plans and incorporate visual servoing for peg grasping and initial alignment.

\bibliographystyle{IEEEtran}
\bibliography{IEEEexample}

\end{document}